\documentclass[10pt,twocolumn,letterpaper]{article}
\usepackage{config/cvpr}     

\usepackage[dvipsnames]{xcolor}

\usepackage{times}
\usepackage{epsfig}
\usepackage{graphicx}
\usepackage{amsmath}
\usepackage{amssymb}
\usepackage{booktabs, nicematrix}
\usepackage{makecell}
\usepackage{multirow}
\usepackage{verbatim}
\usepackage{epsfig}
\usepackage{float}
\usepackage{wrapfig}
\usepackage[accsupp]{axessibility}  
\usepackage[hyphens]{url}
\usepackage{colortbl}
\usepackage{cuted}
\usepackage{bm}
\usepackage{xspace}
\usepackage{multirow}
\usepackage{tabularx}
\usepackage{xcolor}

\usepackage{pifont} 
\usepackage[normalem]{ulem}
\usepackage{arydshln}
\definecolor{cvprblue}{rgb}{0.21,0.49,0.74}

\usepackage[pagebackref,breaklinks,colorlinks,citecolor=cvprblue]{hyperref}

\usepackage{listings}
\definecolor{backcolour}{rgb}{0.95,0.95,0.92}
\lstdefinestyle{mystyle}{
    backgroundcolor=\color{backcolour},  
    keywordstyle=\scriptsize,
    basicstyle=\ttfamily\scriptsize,
    breakatwhitespace=false,
    captionpos=b,                    
    keepspaces=true,                 
    numbers=left,                    
    numbersep=5pt,  
    showspaces=false,                
    showstringspaces=false,
    showtabs=false,                  
    tabsize=4
}

\title{WANDR: Intention-guided Human Motion Generation}
\newcommand{\websiteURL}{\mbox{\href{https://wandr.is.tue.mpg.de}{wandr.is.tue.mpg.de}}}

\newcommand{\methodname}{WANDR\xspace}

\newcommand{\qheading}[1]{\noindent\textbf{#1:}}

\newcommand{\zheading}[1]{\textbf{#1:}}

\newcommand{\HState}{p^{dyn}}  
\newcommand{\HPose}{p}  
\newcommand{\HTrans}{t}  
\newcommand{\HOrient}{r}  
\newcommand{\PDelta}{d}  
\newcommand{\TDeltaz}{\PDelta^{\HTrans_{-z}}}  
\newcommand{\ODeltaz}{\PDelta^{\HOrient_{-z}}}  
\newcommand{\Condition}{c}  
\newcommand{\Intention}{I}  
\newcommand{\IntentionW}{\Intention^{w}}  
\newcommand{\IntentionP}{\Intention^{p}}  
\newcommand{\IntentionR}{\Intention^{r}}  
\newcommand{\Goal}{G}  
\newcommand{\Wrist}{W}  
\newcommand{\Pelvis}{P}  
\newcommand{\Heading}{H}  
\newcommand{\tgoal}{t_{\Goal}}  

\newcommand{\smplx}{\mbox{SMPL-X}\xspace}

\definecolor{GreenColor}{rgb}{0.137,0.573,0.565}
\definecolor{OrangeColor}{rgb}{0.914,0.541,0.0.141}
\definecolor{PurpleColor}{rgb}{0.5,0,0.7}
\definecolor{BlueColor}{rgb}{0,0,1}

\newcommand{\supmatCOLOR}{black}

\newcommand{\supmat}{\textcolor{\supmatCOLOR}{{Sup.~Mat.}}\xspace}

\begin{document}
\author{
Markos Diomataris $^{1,2}$\quad
Nikos Athanasiou$^{1}$\quad
Omid Taheri$^{1}$\quad
Xi Wang$^{2}$\quad
\\
Otmar Hilliges$^{2}$\quad
Michael J. Black$^{1}$\quad \\
\small
$^{1}$Max Planck Institute for Intelligent Systems, T{\"u}bingen, Germany \hspace{0.2in} 
$^{2}$ETH Z\"{u}rich, Switzerland\\
}



\twocolumn[{%
  \renewcommand\twocolumn[1][]{#1}%
 \maketitle
  \vspace*{-1.2cm}
   \begin{center}
    \centerline{ \includegraphics[trim=0 0mm 0mm 0mm, clip=true, width=1. \linewidth]{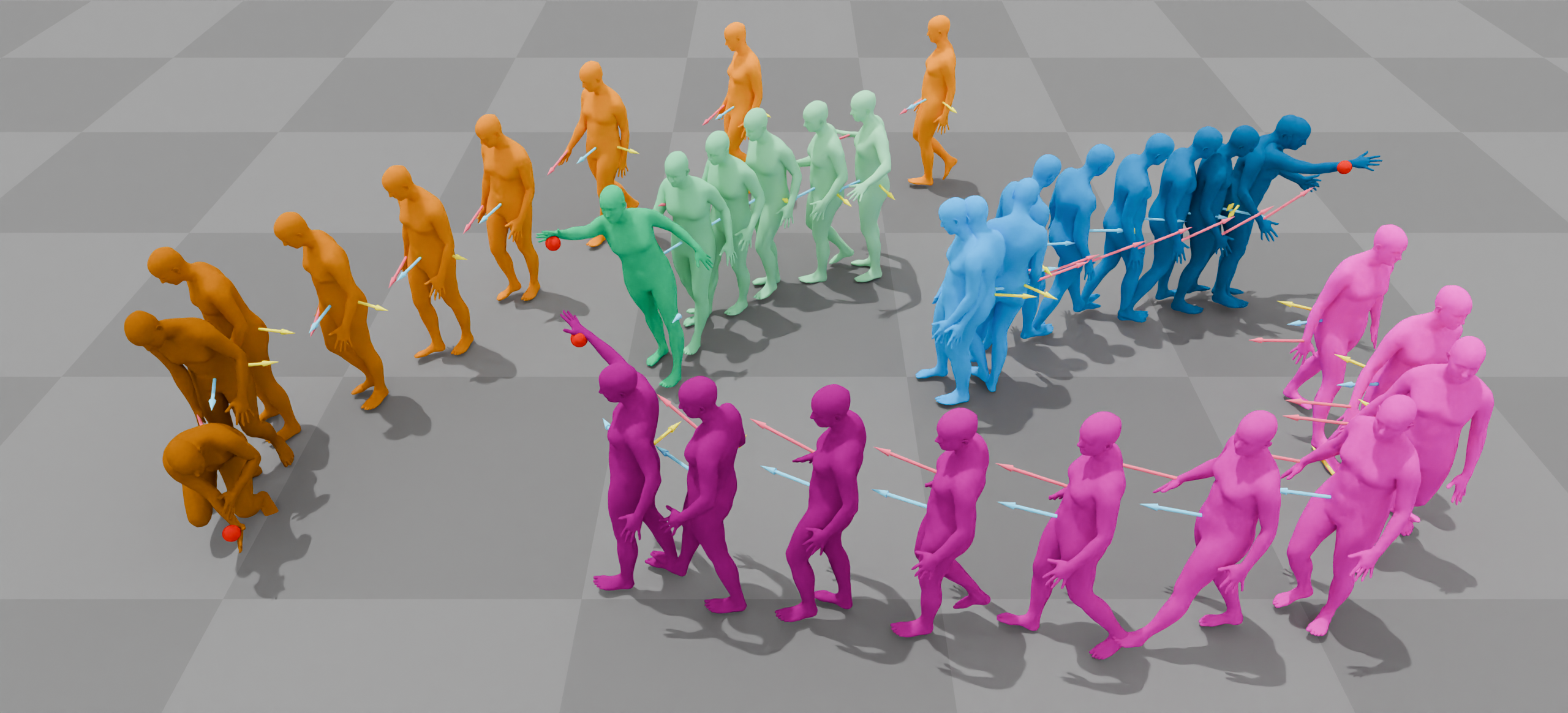}}
 \vspace*{-2.6em}
  \end{center}
  
  \begin{center}
\captionof{figure}{\methodname starts from an arbitrary body pose and generates precise and realistic human motions that reach a specified 3D goal (depicted as a red sphere).
Employing a purely data-driven approach, \methodname is a conditional Variational Autoencoder guided by \textit{intention} features (depicted arrows) that steer the human's orientation (yellow), position (cyan) and wrist (pink) towards the goal.
WANDR is able to reach a wide range of goals even if they deviate significantly from the training data.
}
\label{fig:teaser}
\end{center}%
}]

\begin{abstract}

Synthesizing natural human motions that enable a 3D human avatar to walk and reach for arbitrary goals in 3D space remains an unsolved problem with many applications.
Existing methods (data-driven or using reinforcement learning) are limited in terms of generalization and motion naturalness.
A primary obstacle is the scarcity of training data that combines locomotion with goal reaching.
To address this, we introduce \methodname, a data-driven model that takes an avatar's initial pose and a goal's 3D position and generates natural human motions that place the end effector (wrist) on the goal location.
To solve this, we introduce novel \textit{intention} features that drive rich goal-oriented movement.
\textit{Intention} guides the agent to the goal, and interactively adapts the generation to novel situations without needing to define sub-goals or the entire motion path.  
Crucially, intention allows training on datasets that have goal-oriented motions as well as those that do not.
\methodname is a conditional Variational Auto-Encoder (c-VAE), which we train using the AMASS and CIRCLE datasets.
We evaluate our method extensively and demonstrate its ability to generate natural and long-term motions that reach 3D goals and generalize to unseen goal locations.
Our models and code are available for research purposes at \websiteURL.
\end{abstract}

\vspace{-3mm}
\section{Introduction}
\label{sec:introduction}

Goals drive our motions.
Even the simplest goal can give rise to intricate motions.
Consider reaching for a coffee cup -- it can be as straightforward as an arm extension or can involve the coordinated action of our entire body.
Actions like bending down, extending our arm, and walking must come together to achieve the goal.
At a granular level, we continuously make subtle adjustments to maintain balance and stay on course towards our objective.
The result is a fluid motion that seamlessly integrates numerous smaller movements, all converging toward a common and simple goal: placing our hand on the cup.
Generating this hierarchy of motions,
from the overarching goal to the moment-to-moment individual actions, remains a longstanding challenge in computer vision, graphics, and robotics.

Here we focus on a representative task, illustrated in Fig.~\ref{fig:teaser}: given a goal location in space and a starting pose, a humanoid agent must place an end effector (wrist joint) on the goal location while moving in a natural human-like way.
To solve the task, the agent needs to be able to approach the goal, orient itself towards it, and reach out  such that its wrist makes contact with the goal.
Our primary emphasis is on ensuring autonomy for human agents. Consequently, we strive to minimize the guidance information provided, limiting it only to the human's initial pose and the goal's position.
Diverging from prior data-driven approaches \cite{araujo2023circle, loper2014mosh}, we choose to refrain from evaluating the model solely on limited labeled data. Instead, we devise an evaluation pipeline that requires agents to reach goals positioned in diverse locations around them.
Considering the arbitrary selection of the goal during evaluation, and the minimal guidance information provided, tackling this task is challenging, demanding an approach with the capacity to generalize beyond the distribution of the training dataset.

Existing methods approach this problem either using reinforcement learning (RL)~\cite{Ling2020movaes, Peng2021-vr, HassanInteraction23, Zhang_2022_CVPR} or by capturing task-specific datasets~\cite{taheri2021goal, fan2023arctic, araujo2023circle}.
While RL provides a principled way to explore the solution space, it comes with considerable shortcomings.
The ``trial and error" of exploratory learning, in combination with the high dimensionality of human motion result in policies requiring an enormous amount of training even to achieve simple tasks such as walking to a waypoint~\cite{Zhang_2022_CVPR, braun2023physically}.
In addition, since motion naturalness is better captured by data and not reward functions, RL approaches tend to produce motions that lack naturalness and expressiveness.
Data-driven approaches on the other hand, rely on plentiful training motions that are acquired through motion capture and carefully curated for the downstream tasks \cite{Hassan2021-ak, araujo2023circle}.
Such approaches do not scale and do not generalize well to out-of-distribution tasks.

In prioritizing both motion realism and training efficiency, we adopt a data-driven approach.
However, current data-driven methods lack the ability to learn both from smaller datasets that provide high-quality human reaching motions with goal labels, and from unlabeled larger scale datasets that contain necessary motion skills such as navigating to a goal position.
This raises two key challenges.
First, how do we model human motion in a way that generated motions can combine skills from different datasets?
Second, what should the training objective be in the cases where goal labels are absent?

To address these challenges, we propose \methodname (Wrist-driven Autonomous Navigation for Data-based goal Reaching). 
We observe that by modeling human motion generation as an autoregressive stochastic process that produces motions frame by frame, \methodname is able to combine pieces of different dataset distributions when generating a motion sequence.
Each generation step is conditioned on goal-related information that we call \textit{intention} (visualized arrows in Fig.~\ref{fig:teaser}).
We carefully design \textit{intention} in a way that strikes a balance between being informative enough to guide the avatar to reach the goal, while also being abstract enough to promote generalization to unseen goals.
This allows our generated motions to reach goals that were never encountered during the training phase in a completely zero-shot evaluation scenario.
By generating the motion in an autoregressive way, we disentangle the spatial and temporal dimensions of motion.
This is necessary as it allows our model to generate novel long-term sequences while being realistic in terms of local dynamic details.

In more detail, our method is based on a conditional Variational Auto-Encoder (c-VAE)  that learns to model motion as a frame-by-frame generation process by auto-encoding the pose difference between two adjacent frames.
The condition signal consists of the human's current pose and dynamics along with the \textit{intention} information.
\textit{Intention} is a function of both the current pose and the goal location and therefore actively guides the avatar during the motion generation in a closed loop manner.
Through training, the c-VAE learns the distribution of potential subsequent poses conditioned on the current dynamic state of the human and its \textit{intention} towards a specific goal.
We train \methodname using two datasets: AMASS \cite{Mahmood2019-bi}, which captures a wide range of motions including locomotion, and CIRCLE \cite{araujo2023circle}, which captures reaching motions.

Although AMASS is large, it lacks any explicit label of goals or intentions.
To address this, inspired by the Hindsight Experience Replay paradigm in robotics \cite{HER}, we define \textit{intention} using a hallucinated goal derived from the ground-truth wrist position in a future frame.
This approach allows us to establish a unified training objective spanning AMASS and CIRCLE.
Consequently, our model learns to combine motions from both datasets, enabling it to effectively reach arbitrary goals during testing.

In summary, we present \methodname, a data-driven method that combines an autoregressive motion prior with a novel \textit{intention} guiding mechanism and is able to generate avatars that realistically move in space and reach arbitrary goals.
We experimentally evaluate our approach, including the benefit of combining multiple datasets as well as the generalization capabilities of our motion generator.
Our results underscore the efficacy of the intention mechanism as an elegant way of  guiding the motion generation process while also enabling the incorporation of pseudo goal labels for datasets lacking explicit goal annotations.
The model and code are available for research purposes.
\section{Related Work}
\label{sec:related}
Early research in motion generation focuses on tasks like motion prediction~\cite{ghosh2017learning, aliakbarian2020stochastic, aksan2021spatio, cao2020long, Martinez2017-cu, shu2021spatiotemporal} and unconstrained motion generation~\cite{wang2020adversarial, yu2020structure, wang2020learning, pavlovic2000learning, sigal2010humaneva, ormoneit2000learning, sidenbladh2000stochastic, urtasun2006temporal, cai2021unified, ling2020character}.  
More recently, significant effort has been devoted into improving controllability, with a focus on motion generation conditioned on different types of goals~\cite{xu2023actformer, petrovich2021action}, enabling interactions with scenes~\cite{Hasson2019-ze, huang2022capturing, mir23origin} and objects~\cite{GRAB:2020,  zhang2022couch,taheri2023grip, zheng2022gimo}.
Methods that attempt goal-driven motion generation can be broadly divided into reinforcement learning or data-driven approaches.

\subsection{Reinforcement Learning for Motion Synthesis}

Many existing works employ Reinforcement Learning (RL) for the generation of task-specific long motion sequences. 
Representative work includes MotionVAE~\cite{Ling2020movaes} and AMP~\cite{Peng2021-vr}.
MotionVAE~\cite{Ling2020movaes} employs a two-step process where it initially leverages an autoregressive conditional Variational Autoencoder (VAE) to construct a latent space that encapsulates possible human movements. 
Subsequently, it utilizes RL to sample from this action space to reach a designated target location while avoiding obstacles by monitoring the area ahead. 
Similar to MotionVAE, GAMMA~\cite{Zhang_2022_CVPR} learns a policy to extract samples from a latent space and then employs a tree-based search algorithm to find viable motions that steer clear of obstacles by considering the environmental geometric constraints. 
DIMOS~\cite{Zhao2023Synthesizing} further extends the GAMMA framework by introducing two specialized policy networks: one for locomotion and one for interaction. Together these networks generate goal-conditioned motion sequences that dynamically interact with objects and the environment.  
AMP~\cite{Peng2021-vr} learns an adversarial motion prior from unstructured datasets and then applies goal-conditioned RL. This approach involves the formulation of a style reward to encourage the resemblance of the generated sequences to those in the dataset, complemented by a task-specific reward aimed at achieving a particular objective. 
\citet{HassanInteraction23} extend AMP to produce motions that facilitate interactions with the scene, by conditioning both the discriminator and the policy network on the scene context. 
However, RL requires significant computation and struggles to generate natural and expressive motion sequences. 

\subsection{Data-driven Approaches for Motion Synthesis}
Most data-driven approaches use existing motion capture (MoCap) datasets~\cite{Mahmood2019-bi, trumble2017total} to train their models through supervised learning. 
The pioneering Neural State Machine method~\cite{Starke2019-bl} is a data-driven technique for generating motion with character-scene interactions, focusing on scenarios with a limited number of objects and interactions. 
HuMoR~\cite{rempe2021humor} proposes a robust model for 3D human shape and temporal pose estimation, yet it falls short of generating motions that are conditioned on specific goals. 
The SAMP~\cite{Hassan2021-ak} method, designed for real-time stochastic motion synthesis, generates diverse human-scene interaction movements by breaking down the process into predicting goals, planning paths, and generating motion along a predefined route. 
The GOAL~\cite{taheri2021goal} method, trained on the GRAB dataset~\cite{GRAB:2020}, produces motion sequences in which humans walk towards and grasp 3D objects. However, the generated motions exhibit minimal movements, especially in the feet. 
To address these constraints, the newly introduced CIRCLE~\cite{araujo2023circle} dataset provides a collection of reaching motion data. This dataset is used to train a neural network that generates diverse scene-aware reaching motions. 
Recently, diffusion models have seen the most success at generating motions conditioned on textural input~\cite{tevet2023human, zhang2022motiondiffuse} and spatial data~\cite{karunratanakul2023gmd}. This advancement enables the synthesized motion to accurately reach specified target locations or navigate around obstacles.  
Nevertheless, the effectiveness of data-driven approaches is constrained by the amount of training data and they lack generalization to out-of-distribution scenarios. 



\section{Method}
\label{sec:method}

\begin{figure*}[hbt]
    \centering
    \includegraphics[width=0.9\textwidth, trim=0cm 0.0cm 0cm 0cm, clip=true]{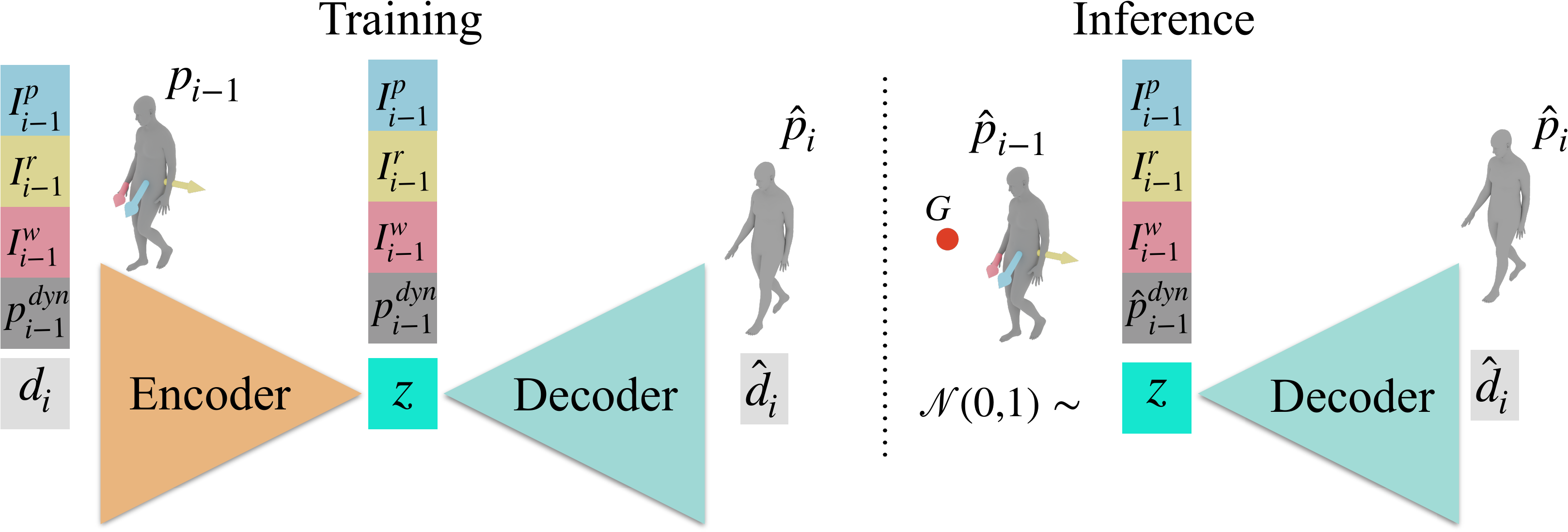}
    \caption{\textbf{\methodname architecture.}
    During training, our model conditions on the intention vectors $I^p, I^r$ and $I^w$, learning to associate them with actions that result into reaching goals realistically.  
    When the training data has no defined goal, we create a goal based on the wrist location in future frames; see Sec.~\ref{sec:intention}.
    The state of the avatar, $\HState_i$ expresses the SMPL-X local pose parameters $\HPose_{i}$, as well as the deltas $\PDelta_{i-1}$ the body parameters have in frame $i-1$. 
    During inference, \methodname takes the intention features, the state, and random noise and returns the change in pose, $\hat{d}_i$.
    The next pose, $\hat{\HPose}_i$ is obtained by integrating the $\hat{d}_i$ with the previous pose $\hat{\HPose}_{i-1}$.}
    \label{fig:model}
\end{figure*}
\begin{figure}
    \centering
    \includegraphics[width=0.45\textwidth, trim=0cm 0cm 0cm 0cm]{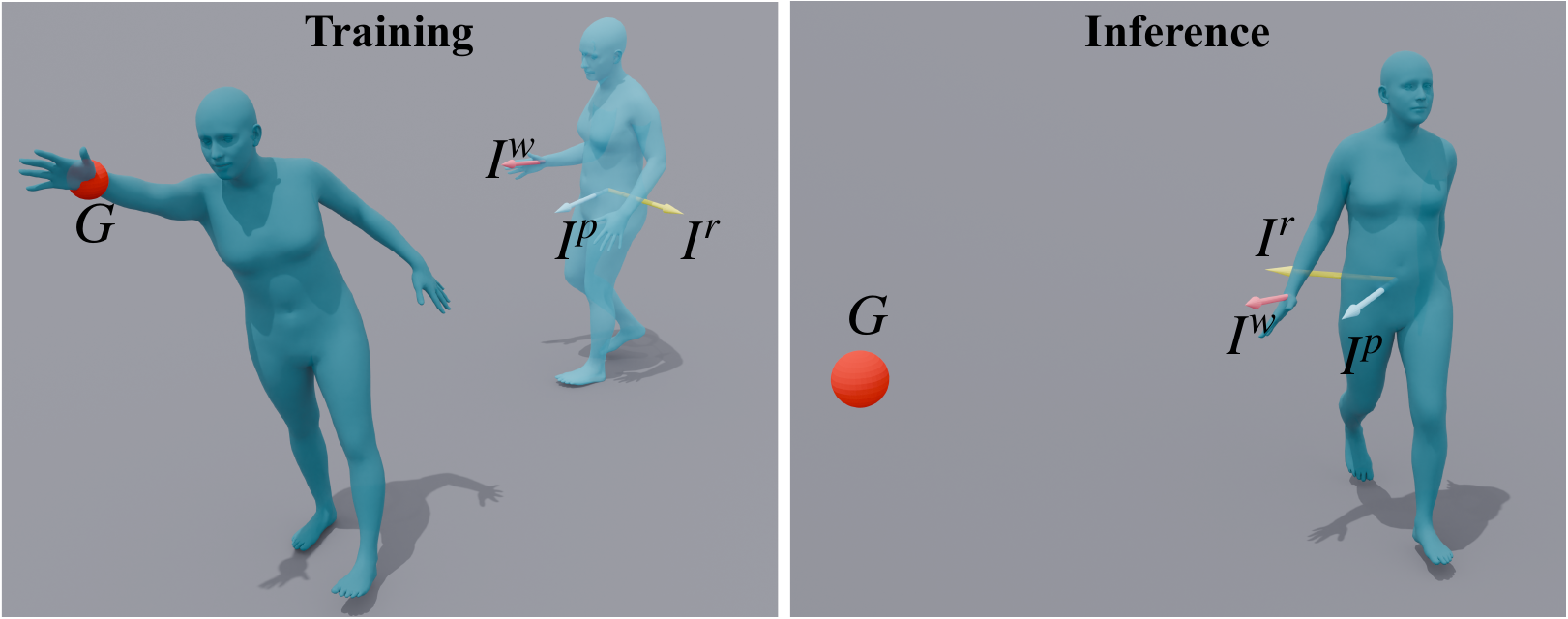}
    \caption{In training, if goals are not specified, they are determined by the future wrist location at a randomly selected future timestep, compensating for the lack of paired ground-truth data in AMASS and direct human motion through intention vectors. During inference, target locations are used as goals with intention vectors calculated based on these specific locations.}
    \label{fig:intention_features}
\end{figure}

Our goal is to have a virtual human that can autonomously and realistically move from an initial pose to an arbitrary goal position and accurately place its right hand on the target. This challenge requires a nuanced understanding of human motion and the intricate dynamics involved in goal-oriented motions. For example, when the human tries to reach a distant goal, the motions are mostly focused on the legs and navigating the body to approach the object, but when it gets close to the object, the focus will be on moving the arms and upper body to reach the target. 
Using these observations, we develop a method named \methodname, which, although trained in a supervised setting on motion capture data, exhibits generalization in reaching unseen goal locations during test time.
\methodname is designed to generate human motion in an autoregressive frame-by-frame fashion, conditioned on novel \textit{intention} features. During training, these features are extracted by picking a future frame as the goal for the wrist. During inference, the \textit{intention} features are dynamically computed based on the goal's position in a feedback loop, guiding the virtual human to reach the goal. See \cref{fig:model} for the network overview.

In this section, we first consider the different distributions of the datasets we will be using (section~\ref{sec:data}).
Following this, we detail the components of the \textit{intention} features and how they are computed during both the training and inference phases (Section~\ref{sec:intention}).
Finally, we define the motion representation and motion generator network (Section~\ref{sec:network}).

\subsection{Two Complementing Datasets} \label{sec:data}

For the development of \methodname, we use two key datasets: AMASS and CIRCLE.
AMASS is a large-scale dataset that offers a broad range of general human motions but lacks a specific focus on goal-reaching tasks. Its diverse collection of movements provides a solid base for understanding human locomotion and body movement when the person is far away from the goal location. In contrast to AMASS, CIRCLE is tailored towards movements involving reaching specific target positions, particularly capturing the nuances of upper body and arm movements. By integrating AMASS's general motion diversity with CIRCLE's targeted goal-reaching data, we equip \methodname with the ability to generate the entire process of reaching a distant goal, from the initial navigation to the precise target-reaching actions.

\subsection{Intention Features}  \label{sec:intention}

We represent 3D human motion as a sequence of \smplx~\cite{smplifyPP} body poses $\textbf{\HPose} = \{\HPose_1, ..., \HPose_N\}$. Each pose $\HPose_i \in \mathbb{R}^{135}$ consists of three concatenated components: the body's translation $\HTrans_{i} \in \mathbb{R}^3$, root orientation $\HOrient_{i} \in \mathbb{R}^6$, and the body pose $\theta_i \in \mathbb{R}^{21 \times 6}$, both in 6D format \cite{Zhou_2019_CVPR}.

For the avatar to reach the goal, it is important to be informed about the spatial relation of the goal location with respect to its current pose, as well as to have a sense of time to reach the goal promptly. We achieve this by introducing the \textit{intention} features, which are central to our approach.

To define the \textit{intention} features \(\Intention_{i}\) at timestep $i$, it is crucial to first establish the selection criteria for the goal $\Goal\ \in \mathbb{R}^3$, within a motion sequence. Our training involves both label-available scenarios (e.g., CIRCLE dataset) and label-absent scenarios (e.g., AMASS dataset). In label-available cases, where the goal position \(\Goal\) and the frame index \(\tgoal\) which it is reached are known, calculating $\Intention_{i}$ is straightforward. Conversely, for label-absent scenarios like in AMASS, we pick a random future frame as the $\tgoal$ and define the human's right wrist joint location as the goal $\Goal$ (Fig.~\ref{fig:intention_features}).

In both scenarios, the \textit{intention} features are defined as:
$$
\Intention_{i} = 
    \Intention_{i}(\HPose_{i}, \Goal, \tgoal, i).
$$
These features are essentially a function of the current body pose, the goal position, and time, offering both spatial and temporal insights required for the motion to reach the goal.
They are designed to provide sufficient information to reach goals while also enabling test-time generalization.
We define them as three distinct components:
$$\Intention_{i} = (\IntentionW_{i}, \IntentionR_{i}, \IntentionP_{i}).$$
These components represent wrist intention, body orientation intention, and pelvis intention, respectively. 

\qheading{Wrist Intention} This is the main time-dependent component that guides the wrist to reach the goal. It is calculated as the necessary average velocity for the wrist to be at the goal location in time, defined as:
$$\IntentionW_{i} = 
\frac{\Goal-\Wrist_{i}}{\tgoal - i}
$$
where $\tgoal$ is the frame when the goal should be reached. During training, we know the goal frame and the time to reach it. At test time, we know the goal location but need an externally-supplied time to reach it.
An emergent behavior of this formulation is that, at inference, the model is able to adjust its movement speed and reach the goal just in time.

\qheading{Orientation Intention} This component captures the body orientation when reaching the goal location. By conditioning on this, we ensure that the human model orients towards the goal and smoothly navigates towards it, preventing unnatural motions during inference, like walking backward.  During training, this is defined as the difference between the forward direction of the current body frame, $\Heading^{xy}_{i}$, and the goal body, $H_{\tgoal}^{xy}$ where $xy$ signifies removing the $z$ component. During inference, since we do not have the goal body, we use the pelvis position $\Pelvis_{i}$ to calculate the pelvis-to-goal direction as the desired orientation. This feature is formulated as:
$$\IntentionR_{i} = \begin{cases}
H_{\tgoal}^{xy} - \Heading^{xy}_{i} & \text{during training}\\
(\Goal - \Pelvis_{i})^{xy} - \Heading^{xy}_{i} & \text{during inference}.\\
\end{cases}$$

\qheading{Pelvis Intention} This feature captures information about the position of the goal relative to the body.
It is the difference between the goal and the pelvis joint, excluding the z (height) component. 
Following the approach in \cite{taheri2021goal}, we scale this distance by an exponential function that saturates this vector to have a maximum norm of $2$. This formulation helps the method generalize to navigating towards the goal during longer motions and helps the model learn since the distance from the goal does not grow indefinitely in extreme scenarios.
This intention is defined by the equation:
$$\IntentionP_{i} = 2 \times (1 - e^{||\Goal^{xy}-\Pelvis^{xy}_{i}||_{2}}) \times \frac{\Goal^{xy}-\Pelvis^{xy}_{i}}{||\Goal^{xy}-\Pelvis^{xy}_{i}||_{2}}.$$

In the experimental section, we delve into the significance of each of these intention features and discuss the rationale behind our design choices, illustrating their impact on the effectiveness of our model.

\subsection{Motion Network (\methodname)} \label{sec:network}

\methodname is designed as a conditional Variational Auto-Encoder (c-VAE) network, operating in an autoregressive manner to generate sequential motion frames. This framework is pivotal in predicting the subsequent pose in a motion sequence, emphasizing an incremental, frame-by-frame approach.

Central to our approach is the training of the c-VAE to autoencode pose deltas, denoted as $\PDelta_i \in \mathbb{R}^{135}$. These deltas represent the difference between two consecutive poses, $\HPose_{i}$ and $\HPose_{i-1}$. By focusing on pose deltas rather than absolute pose values, our model benefits from an important inductive bias, enhancing its learning efficiency and performance, as supported by prior research \cite{rempe2021humor, Ling2020movaes}.
We separate rotational differences into: body orientation ($\PDelta^{\HOrient}_{i}$) and body pose ($\PDelta^{\theta}_{i}$), each expressed in a 6-D rotational format. Translation deltas are denoted as $\PDelta^{t}_{i} = \HTrans_{i} - \HTrans_{i-1}$.

To enhance the motion representation's invariance, we remove information related to the global z-orientation. This is accomplished by subtracting the global z Euler angle of $\HOrient_{i-1}$ from both the translational ($\PDelta^{t}_{i}$) and rotational ($\PDelta^{r}_{i}$) deltas. The resulting deltas, $\TDeltaz_{i}$ for translation and $\ODeltaz_{i}$ for orientation, provide a more robust and consistent representation of motion, irrespective of global direction. Consequently, the delta pose features for any given frame $i$ are composed as follows:
$$\PDelta_{i} = (\TDeltaz_{i}, \ODeltaz_{i}, \PDelta^{\theta}_{i}).$$
An advantage of this representation is its consistency across different motion global orientations. For instance, in the scenario of a person walking, the delta representation remains agnostic to the walking direction. This attribute underscores the efficacy of our method in capturing the essence of motion without being biased towards any specific orientation or direction.

\qheading{Condition Inputs}
For each motion frame, the decoder is conditioned on a combination of state and intention features. Specifically, this condition signal is formulated as $\Condition_{i} = (\HState_{i}, \Intention_{i})$. The state features, $\HState_{i}$, encapsulate the avatar's current local pose, focusing on the z-component of translation and a modified orientation that excludes the global z Euler angle, along with the pose deltas $\PDelta_{i-1}$ of the previous step. That combination ensures that the generated motion at each step is informed by both the local pose configuration of the avatar, its dynamics and its directional intention towards the set goal, vital for producing realistic, goal-oriented human motions.
An overview of the network architecture is shown in Fig.~\ref{fig:model}.

\subsection{Training Losses}

Our training objective is a composite of three distinct loss functions:
$$\mathcal{L} = \mathcal{L}_{rec} + \alpha \mathcal{L}_{KL} + \mathcal{L}_{J}.$$
The reconstruction loss, $\mathcal{L}_{rec}$, measures the accuracy of the motion reconstruction, quantified as the mean square error (MSE) between the input pose delta, $\PDelta_{i}$, and its reconstructed counterpart, $\hat{\PDelta}_{i}$. It ensures the network's ability to faithfully replicate the input motion.

The KL Divergence Loss, ($\mathcal{L}_{KL}$), evaluates the deviation of the encoded distribution from a standard normal distribution. It is formulated as:
$$\mathcal{L}_{KL} = \mathcal{KL}(\mathcal{N}(0, I)||\mathcal{N}(\mu_{i}, \sigma_{i})).$$
Here, $\mu_{i}$ and $\sigma_{i}$ represent the mean and variance of the Gaussian distribution predicted by the encoder. We balance this term with $\alpha = 10^{-2}$ to prevent the over-dominance of $\mathcal{L}_{KL}$, thereby aiding the decoder in avoiding collapse to mean predictions.

Finally, we use a Joint Error Loss ($\mathcal{L}_{J}$), to ensure perceptual accuracy, by integrating the predicted $\hat{\PDelta}_{i}$ to get the predicted next pose $\hat{\HPose}_{i}$, which is then fed into the SMPL-X model to obtain the predicted joint positions, $\hat{J}$. The loss $\mathcal{L}_{J}$ is the MSE between these predicted joints $\hat{J}$ and the ground truth joints $J$, addressing errors that might not be apparent in parameter space but are perceptually significant, such as incorrect body orientation.

Notably, our approach does not incorporate any explicit loss functions directly related to reaching a goal. This omission is a deliberate choice, aligning with our method's emphasis on generalizing to diverse goal-reaching scenarios without being constrained by goal-specific training losses.

\subsection{Motion Generation}

In the inference phase of \methodname, our primary objective is to generate human motion that is driven towards a specific goal.
Using the decoder of the \methodname c-VAE, we iteratively generate and integrate pose deltas. This process is initiated from the starting pose and progressively builds upon each subsequent pose.

The intention features are recalculated at each step based on the current predicted pose and the goal location. They serve as a guiding mechanism, ensuring that the generated motion is consistently oriented towards placing the human's right wrist on the target.

The user can control the motion's pace by specifying the time to reach the goal $\tgoal$. This directly affects the wrist intention feature, enabling adjustments from fast to slow motions to suit various scenarios and constraints.

\section{Experiments}
\label{sec:evaluation}
In this section, we outline the datasets used for training and evaluation and benchmark how each dataset affects the goal-reaching ability and the quality of the generated motions. Furthermore, we compare our approach with several baselines and ablate the effect of the different components of our \textit{intention} vector.

\subsection{Datasets \& Processing}
Our model is trained on two datasets: AMASS~\cite{Mahmood2019-bi} and CIRCLE~\cite{araujo2023circle}. AMASS is a collection of 17k sequences, containing a wide range of motion types including long-term navigational skills like walking and turning. CIRCLE, on the other hand, contains 7.2k shorter sequences, each marked with a specific goal reached by a hand. For training, we refine AMASS by excluding sequences where feet are more than $20cm$ above the ground, resulting in a combined dataset of nearly 20k sequences. This dataset is split into 80\% training, 10\% validation, and 10\% test sets.
All motions are re-sampled to 30 frames per second (fps).

\subsection{Evaluation Strategy}

Our evaluation procedure aims at testing the degree which \methodname can generalize to generating reaching motions that start from unseen poses and reach the whole range of 3D space around the starting pose. This is why we choose not to evaluate on held-out motion-goal pairs from the training data. Instead, we only hold out initial poses. During evaluation, starting from these unseen poses, we generate motions that attempt to reach goals that uniformly cover the volume of a cylinder centered on the human, including completely out-of-distribution goal locations (see \supmat Sec.~\ref{sec:eval_distribution}).

In particular, the set of evaluation goals is defined in a cylindrical coordinate frame by taking all the combinations of (1) $5$ angles equally separating the 360 degrees around the human, (2) $5$ different goal heights ranging from $0$ to $1.8$ meters and (3) $5$ distances from $0.5$ to $5$ meters. 
We generate motions from $6$ different initial poses, with an $8$-second duration specified for reaching each goal. Five motions are sampled for each pose-goal combination, resulting in $5 \times 5 \times 5 \times 6 \times 5 = 3750$ unique motion sequences from which our metrics are computed.
This setup allows us to thoroughly test our model in diverse scenarios, including long-term movements, navigational skills, and reaching motions at various heights and distances.

\begin{figure*}[!h]
    \centering
    \includegraphics[width=\textwidth]{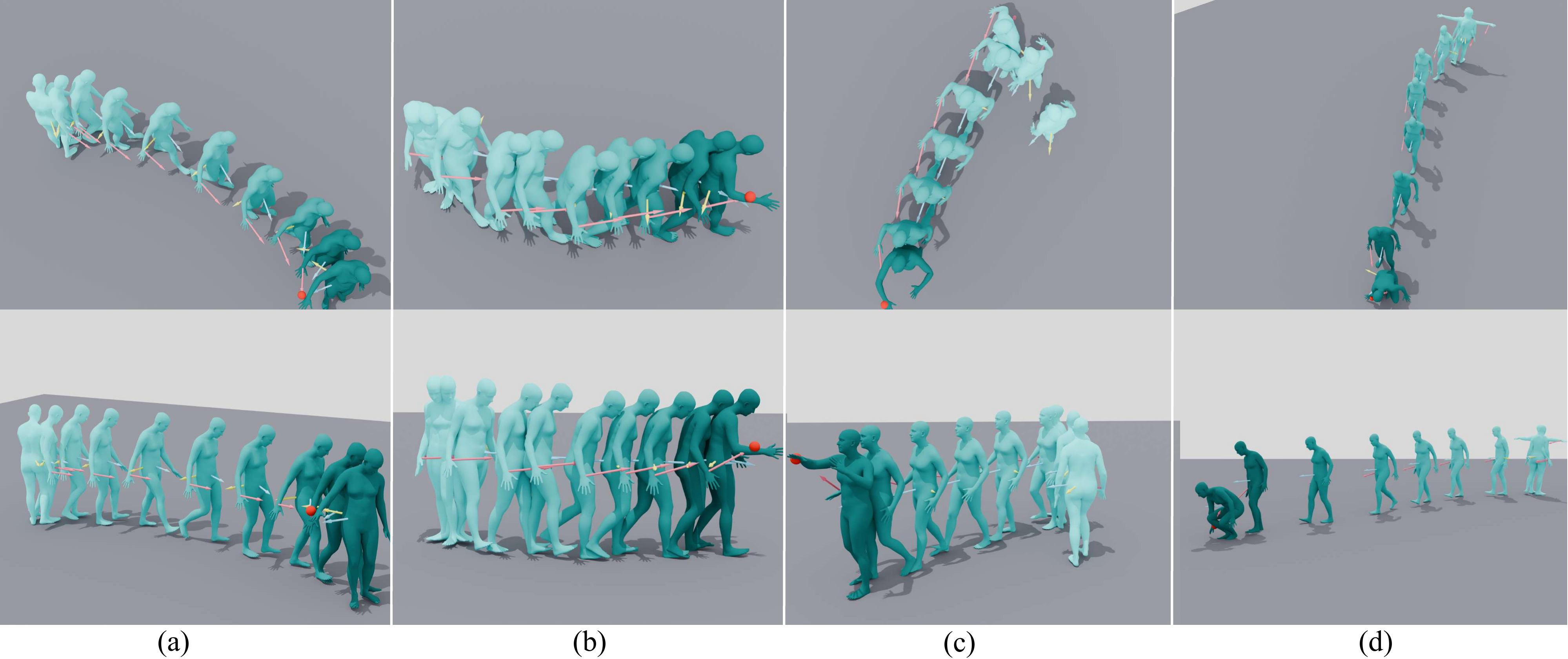}
    \caption{Diverse motion generated with \methodname: Displaying a range of motions generated by \methodname from various initial poses towards arbitrary goals. Examples include navigating towards goals from initial orientations not facing the goal (a, b, c, d), elevating the right hand to reach higher targets (c), and bending down to access goals near the floor (d), showcasing the model's ability to adapt to novel goal locations.}
    \label{fig:quals}
\end{figure*}

\subsection{Evaluation Metrics}
To accurately assess the effectiveness of our approach in generating realistic, goal-oriented human motion, we employ a set of metrics focused on both the ability to successfully reach the intended goal and the naturalness of the motion. These metrics are:

\begin{itemize}
    \item \textbf{Success Rate (SR)}: This quantifies the percentage of motions where the right wrist reaches within $10$cm of the goal, indicating successful goal attainment. The criterion for success aligns with that used in \cite{araujo2023circle}.

    \item \textbf{Foot Skating (FS)}: FS evaluates the naturalness of the motion based on foot skating, where a frame is considered as having foot skating if the lowest vertex of the human mesh moves more than $0.66 cm$ between consecutive frames (adjusted from the $1$cm threshold used in \cite{araujo2023circle} to accommodate our 30fps motion generation).

    \item \textbf{Distance to Goal (DTG)}: DTG records the closest distance in cm that the right wrist gets to the goal during a motion. This metric offers a nuanced view of the model's capability to guide the motion towards the goal.

\end{itemize}

\subsection{Results}
\subsubsection{Quantitative Results}
\qheading{Combining AMASS and CIRCLE}
Our evaluation in Table~\ref{tab:dataset_ablation} validates our hypothesis about the benefits of training with both the AMASS and CIRCLE datasets.
On the one hand, training only on CIRCLE (line 1) is not sufficient for the model to learn necessary navigational skills, such as walking, due to the dataset's narrow focus or reaching motions. This is apparent from the very high foot-skating.
On the other hand, training only on AMASS (line 2) results in high-quality motion generation with low foot-skating, but a relatively low success rate in goal-reaching.
The Distance to Goal (DTG) metric, suggests that the model is able to navigate close to the target, but it lacks the precise movement needed to successfully reach for the goal.
Using both datasets (line 3) illustrates how our approach effectively merges the broad motion vocabulary of AMASS with the goal-oriented precision of CIRCLE, leading to both high-quality motion and improved goal-reaching capability.

We also compare with GOAL~\cite{taheri2021goal}, a method that generates human motions that reach and grasp objects. GOAL is trained on GRAB~\cite{GRAB:2020}, a dataset purely consisting of motions of humans grasping and manipulating objects. Since the relative positioning of the human and the object as well as the motions in GRAB have very small variations, GOAL does not succeed in any of the evaluation configurations (line 4).


\begin{table}[tb]
\centering
\begin{tabular}{l|ccc}
\Xhline{2\arrayrulewidth}
\textbf{Train Set} & \textbf{SR $\uparrow$} & \textbf{FS $\downarrow$} & \textbf{DTG (cm) $\downarrow$} \\
\hline
\methodname/Circle & 0\% & 56\% & 205.4 \\
\methodname/AMASS  & 16\% & 19\% & 48.0 \\
\methodname        & \textbf{32}\% & \textbf{16}\% & \textbf{24.8} \\
\hline
GOAL \cite{taheri2021goal} & 0\% & 29\% & 149.2 \\ 
\Xhline{2\arrayrulewidth}
\end{tabular}
\caption{We evaluate \methodname trained on different datasets and compare with GOAL~\cite{taheri2021goal}. Training solely on CIRCLE results in unrealistic motions, whereas AMASS excels in motion quality but struggles with finer goal-reaching skills. WANDR, leveraging both of what these datasets offer, demonstrates realistic motions as well as better ability to reach goals compared to baselines and existing methods.
}
\vspace{-3mm}
\label{tab:dataset_ablation}
\end{table}

\qheading{Ablation of Intention Features}
In order to demonstrate the contribution of each component of the \textit{intention} feature we conduct an ablation study (Table~\ref{tab:intention-opt}).
Using only wrist intention (line 1) results in the lowest foot skating, due to the minimal constraints applied to the motion, allowing for more adaptable motion planning.
But wrist intention features, are time-dependent and do not carry information about the absolute distance to the goal. This is why it can lead to the avatar over- or under-shooting and thus achieving a low success ratio (SR). 
The addition of pelvis intention (line 2) enables the avatar to sense the distance to the goal while the orientation intention (line 3) properly aligns the body to face the goal.
Since pelvis and orientation intention add more constraints to the motion, they can sometimes cause more challenging body dynamics and produce motions with higher foot-skating (FS) (e.g. turning around in place instead of walking in a U-turn).
We also try removing the \textit{intention} from the motion prior and optimizing the latent space of a randomly generated initial motion to minimize the distance between the wrist and the goal (VAE + opt). This approach fails since the result is heavily dependent on the initialization of the latent variables of the motion.
Our results confirm that each component of the intention feature is essential to achieving the overall performance of the model. 
For more details on the optimization see \supmat

\begin{table}[h]
\centering
\begin{tabular}{lccc}
\Xhline{2\arrayrulewidth} 
\textbf{Train Set} & \textbf{SR $\uparrow$} & \textbf{FS $\downarrow$} & \textbf{DTG (cm) $\downarrow$} \\ \Xhline{2\arrayrulewidth}

\methodname ($\IntentionW$) & 15\% & \textbf{13}\% & 62.9 \\
\methodname ($\IntentionW+\IntentionP$) & 18\% & 17\% & 44.9 \\
\methodname ($\IntentionW+\IntentionR$) & 19\% & 19\% & 36.0 \\
\methodname (full \textit{intention}) & \textbf{32}\% & 16\% & \textbf{24.8} \\
\hline
VAE + opt & 3\% & 4\% & 217.0 \\
\Xhline{2\arrayrulewidth}
\end{tabular}
\caption{\textbf{Ablation Study.} We evaluate the impact of each component of the intention vector.
We also compare with an optimization baseline that does not use any condition signals. 
The results highlight the effectiveness of all of the components of intention as well as the fact that the complexity of the task makes ``brute-forcing'' with optimization unsuccessful.
}
\label{tab:intention-opt}
\end{table}

\begin{figure}
    \centering
    \includegraphics[trim=000mm 000mm 000mm 000mm, clip=true, width=1. \linewidth]{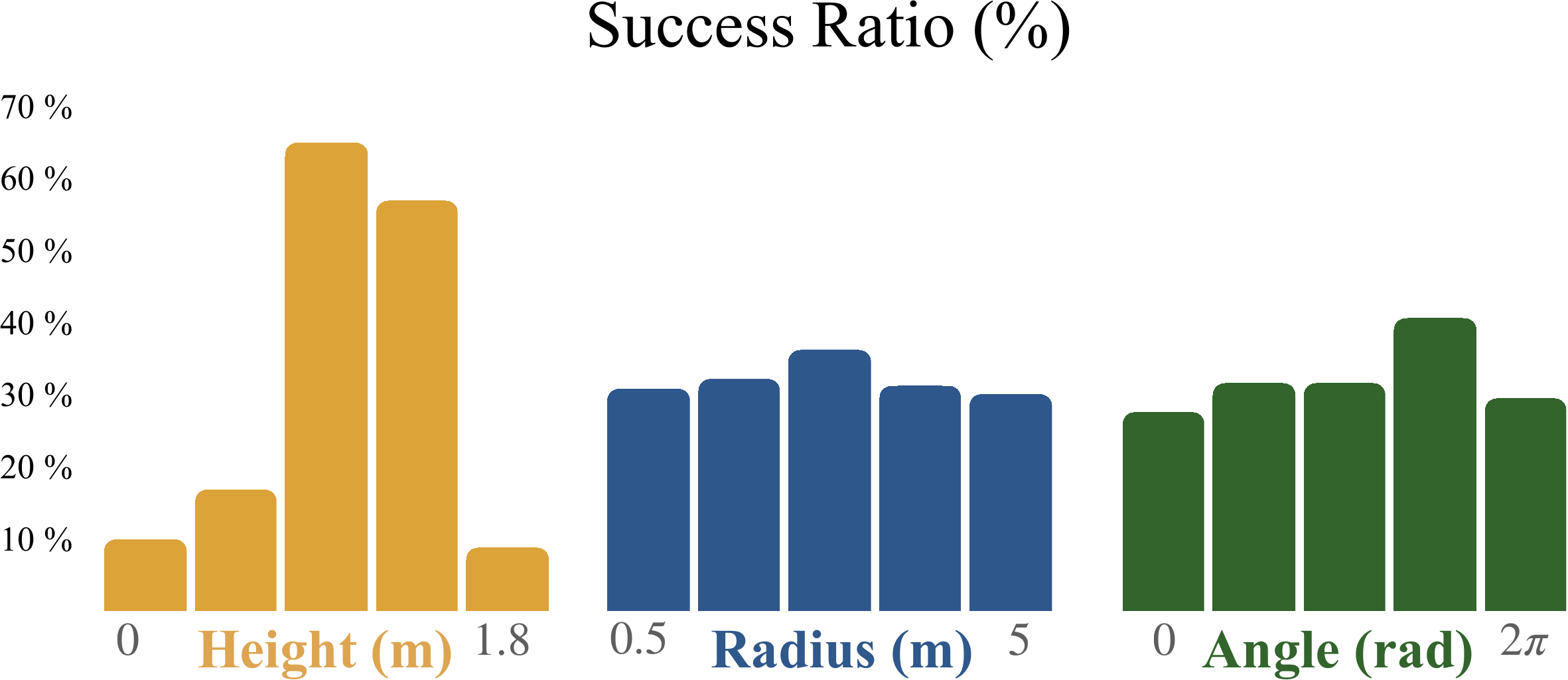}
    \captionof{figure}{We show the success rates of reaching goals at various heights, angles, and distances from the initial human pose. It highlights how goal position affects the model in accurately navigating and achieving the goals.}
    \label{fig:plot_for_goals}
\end{figure}

\qheading{Success Ratio Distribution}
Our decomposition of the model's success ratio, presented in Fig.~\ref{fig:plot_for_goals}, offers insights into how the model's performance varies with respect to different goal positions. \methodname demonstrates a consistent ability to reach goals across various distances (blue) and directions (green). It is more capable at reaching goals that are closer to the natural position of the wrist and do not require extensive bending or stretching (yellow). This trend likely results from the abundance of standing or upright motion sequences in the training data, as opposed to motions involving bending or crouching. This analysis provides valuable information for future improvements and dataset balancing.

\subsubsection{Qualitative Results}
In \cref{fig:quals}, we show a variety of motion sequences generated with our network featuring reaching goals located at varying distances and heights, highlighting the model's ability to realistically and smoothly orient, navigate, and reach for goals. These goals require actions such as bending down, turning, or stretching upwards. A critical aspect observed is the model's ability to decelerate as it approaches the goal, seamlessly coordinating body and arm movements to achieve a natural-looking reaching motion. Overall, the qualitative results show that our network generalizes well to novel goal locations while generating realistic motions. For more results please see \supmat and the video.


\section{Conclusion}

In conclusion, our research presents a novel data-driven approach to human motion generation, focusing on the task of reaching arbitrary goals in space.
We introduce novel \textit{intention} features that enable learning both general navigational skills from AMASS and goal-reaching skills from the CIRCLE dataset under the same distribution.
We evaluate our model's ability to reach unseen goals that cover the whole space an avatar should be able to reach around it.
The autoregressive design of \methodname demonstrates generalizability in generating realistic human motions that reach unseen goals without requiring any extra guidance information such as a pre-defined trajectory.

\zheading{Limitations and Future Work}
Our approach is not without its limitations. Currently, error accumulation can sometimes bring the avatar to states where it can no longer recover. Additionally, our model shows less proficiency in reaching extremely low or high goals, reflecting a need for more diverse training data encompassing a wider range of body movements. Future work could focus on incorporating realistic grasping mechanisms and interactions with objects, as well as including scene navigation capabilities. This could involve integrating more complex datasets or developing advanced algorithms capable of understanding and interacting with varied environmental contexts, thereby pushing the boundaries of realistic human motion simulation.
\newline
\qheading{Acknowledgments}
Markos Diomataris was supported in part by the Max Planck ETH Center for Learning Systems.
We thank
P. Ghosh,
O. Ben Dov
S. Tripathi,
for the fruitful discussions and
A. Cseke,
T. Niewiadomski,
T. Alexiadis,
T. McConnell,
for conducting the user studies.
\newline
\qheading{Conflicts of Interest} 

\href{https://files.is.tue.mpg.de/black/CoI_CVPR_2024.txt}{https://files.is.tue.mpg.de/black/CoI\_CVPR\_2024.txt}

\bigskip

\maketitle
\newpage
\appendix
{\noindent\LARGE\textbf{Supplementary Material}}
\newline
\renewcommand{\thefigure}{S.\arabic{figure}}
\renewcommand{\thetable}{S.\arabic{table}}
\renewcommand{\theequation}{S.\arabic{equation}}
\setcounter{figure}{0}
\setcounter{table}{0}
\setcounter{equation}{0}

\section{Introduction}
\label{sec:introduction}
This supplemental material offers more details regarding the use of our method in an optimization framework, 
the effect of motion duration on the generated motions, different applications of our method, and more qualitative results. 
Please see the \textbf{Supplementary Video}, where we extensively demonstrate the realism  and adaptability of our generated reaching motions across diverse scenarios.

The video effectively contains:
(1)     the problem and our motivation, 
(2)     our method and key ideas, 
(3)     multiple example motions, and
(4)     different applications of our method such as extension to reaching dynamic goals.
The video serves as a dynamic and illustrative supplement, showcasing our contributions in a manner that is hard to show in a paper format.

\section{Technical Implementation}
\label{sec:tech_details}

\subsection{Model Architecture}
\methodname c-VAE architecture~\cite{Kingma2014} employs an Encoder and a Decoder, each composed of fifteen layers in a Multi-Layer Perceptron (MLP) configuration. We integrate relu activation functions, dropout and layer normalization at each stage for enhanced performance. The latent space is represented as a 64-dimensional vector. 
In our design, the condition signal of the c-VAE is concatenated with the input delta (in the case of the Encoder) and the latent vector (in the case of the Decoder).

\subsection{Training Details}
We developed and trained our method using the PyTorch framework~\cite{pytorch}. We train it for 900 epochs on 4 Tesla V100 GPUs. We use a batch size of 512, resulting in approximately 20 hours of training duration. For optimization, we use Adam \cite{kingma2014adam} with a starting learning rate of $1e-4$ that linearly decreases to $1e-5$ during training.

A crucial aspect of our training regimen includes performing a teacher-forcing method, which involves feeding the model's own predictions back into the input.
This process facilitates the Decoder network in acquiring the capability to compensate for potential errors that may arise during the prediction of deltas.
During the whole process of the training, the c-VAE is being trained on the task of auto-encoding the motion deltas.
As the training progresses, we additionally perform motion generation for a few steps.
Specifically, we reconstruct the deltas, integrate them to obtain the subsequent pose, and then sample from the latent space while conditioning on the generated pose. 
We repeat this process for up to $s$ steps increasing the $s$ linearly from $0$ up to 10 along the span of 50 epochs and then keeping it fixed.

\subsection{Optimization Details} \label{sec:optimization}
The formulation of our method as a c-VAE provides us with a smooth latent space that allows us to search this manifold in an optimization process to reach various target goals. For this, we apply specific constraints to the decoder's output, i.e.~the body poses, and optimize the latent space representation of the poses to achieve the desired motions. 

In 
Tab.~\ref{tab:intention-opt}
of the main manuscript, we explore how optimizing a trained motion prior performs compared to our method without any optimization, in achieving various goals. The results show that \methodname without optimization performs better than the other methods, even with optimization.

\begin{figure*}[!h]
    \centering
    \includegraphics[width= 1.\textwidth, trim=0cm 0cm 0cm 0cm]{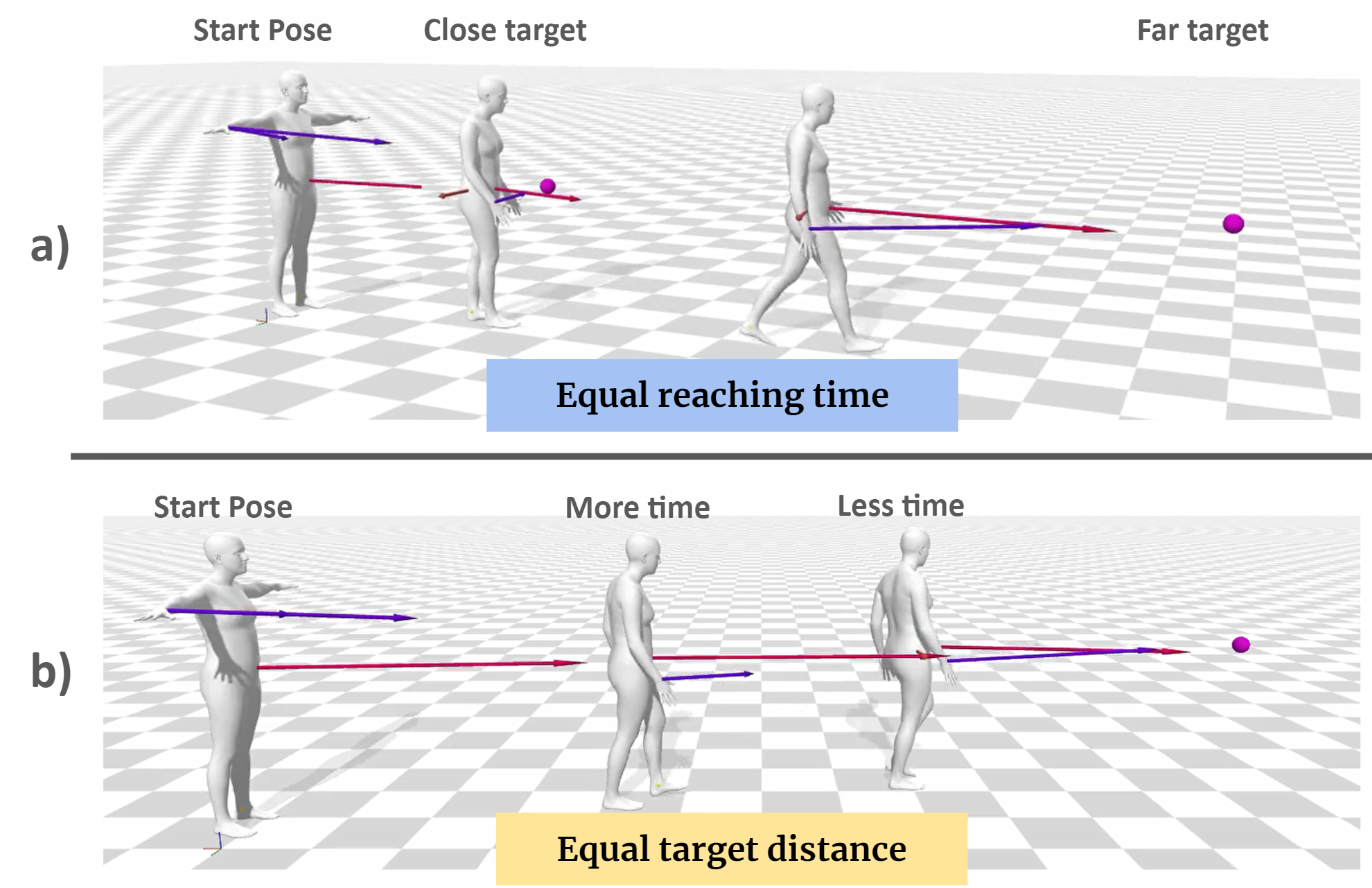}
    \caption{Generated motions for different goals with the same time constraint (a) and for a similar goal with varying time constraints (b). The results show that our dynamic intent features enable the adaptability of \methodname to generate time-controlled motions.}
    \label{fig:time_goal}
\end{figure*}

To do the optimization, we first generate a motion to reach a goal using \methodname, and then refine the motion through optimization aiming to align the wrist's position in the final frame more closely with the target goal.

In order to achieve this, we employ a dual-component loss function:
$$\mathcal{L}_{opt} = \mathcal{L}_{norm} + \mathcal{L}_{goal}$$
Here, $\mathcal{L}_{norm}$ represents the log-likelihood of the motion's latent vectors under a normal distribution. $\mathcal{L}_{goal}$ calculates the mean square error between the wrist's final frame location and the goal. $\mathcal{L}_{norm}$ seeks to maintain the generated motion within plausible human movements, while $\mathcal{L}_{goal}$ specifically tunes the motion to bring the wrist in proximity to the goal in the final frame.
Notably, even though $\mathcal{L}_{goal}$ is applied only on the final frame, its gradient flows across the whole sequence since the autoregressive generation process is fully differentiable.

\cref{sec:applications}  shows our method's ability to produce motions with different time durations, as well as its integration with the optimization framework presented in the main paper.

\section{\methodname Applications}
\label{sec:applications}
In this section we show that our method can be used in various scenarios and for different applications.

\subsection{Time-controlled Motions}
One key aspect of our intention features is the dependency of the wrist-intention vector on the goal-reaching time, enabling the generation of time-controlled motions. As mentioned in the main paper, this vector is computed by dividing the distance from the wrist to the goal by the time remaining to reach it. During inference, by changing the reaching time or distance, the generated motions adapt and become rapid or slow. Figure \ref{fig:time_goal} presents two scenarios: (a) reaching different goals within the same time duration, and (b) reaching a goal within different time durations. These cases illustrate how the motion varies in response to the time and distance parameters. For qualitative examples, please see the \textbf{Supplementary Video}.

\subsection{Optimization-enabled extensions}
\begin{figure*}[!h]
    \centering
    \includegraphics[width= 1. \textwidth, trim=0cm 0cm 0cm 0cm]{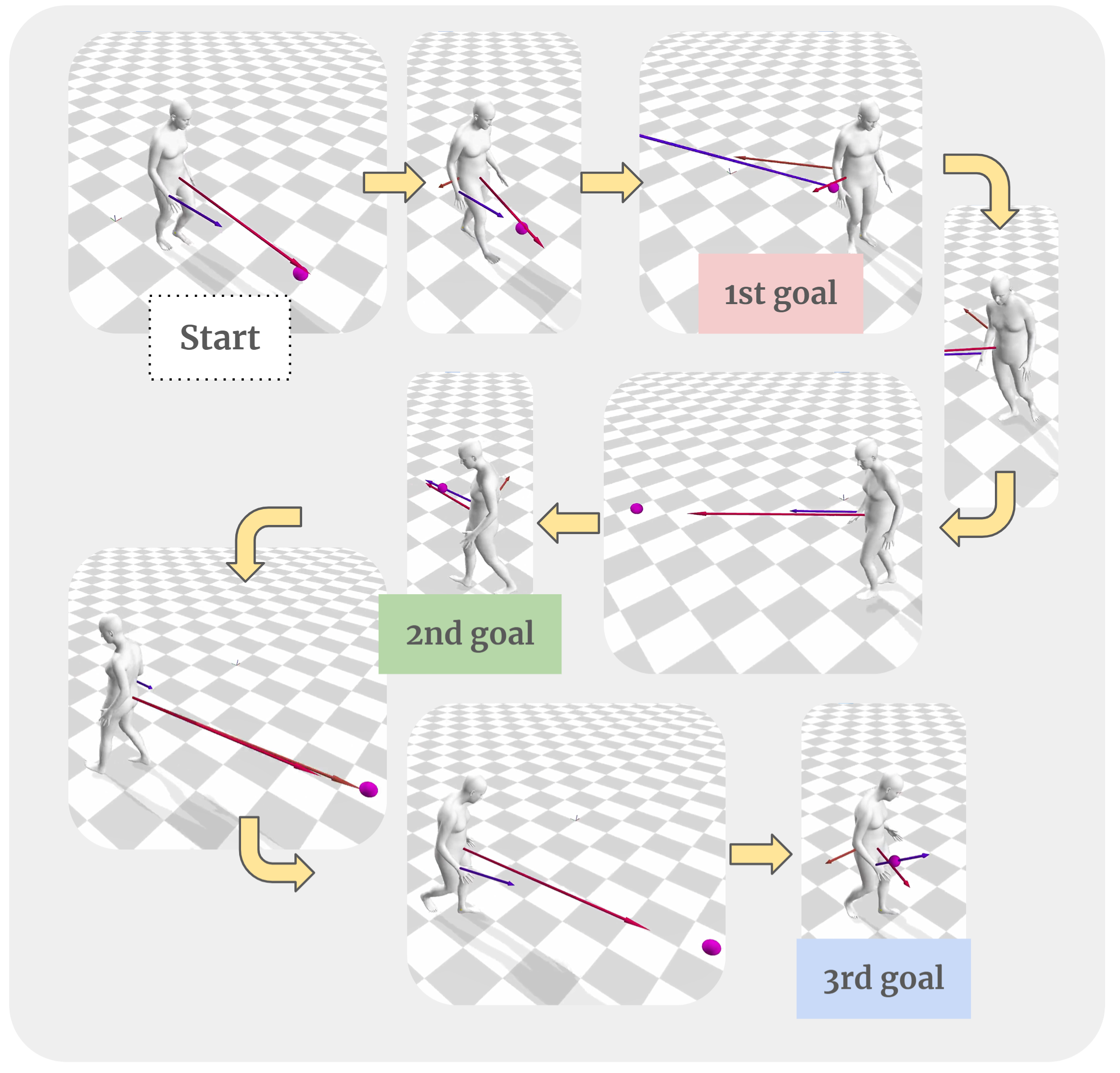}
    \caption{A demonstration of \methodname generating a motion sequence to achieve multiple goals. The intention features are recalculated and updated after each iteration of the autoregressive process, enabling dynamic goal adjustments during the motion generation.}
    \label{fig:multi_goal}
\end{figure*}

\qheading{Multi-goal Reaching}
We show that our unique intention features enable the generation of motions to achieve multiple goals sequentially. Although trained for single-goal achievement, the dynamic nature of our intention features allows for multiple goal definitions during inference. These features are recalculated and updated at each iteration of \methodname's autoregressive process, adapting to changes in goal locations. Figure \ref{fig:multi_goal} illustrates this capability, where a motion sequence is generated to achieve several goals. This adaptability also extends to tracking and following moving targets, as shown in our \textbf{Supplementary Video}.

\begin{figure}
    \centering
    \includegraphics[width= 0.45\textwidth, trim=0cm 0cm 0cm 0cm]{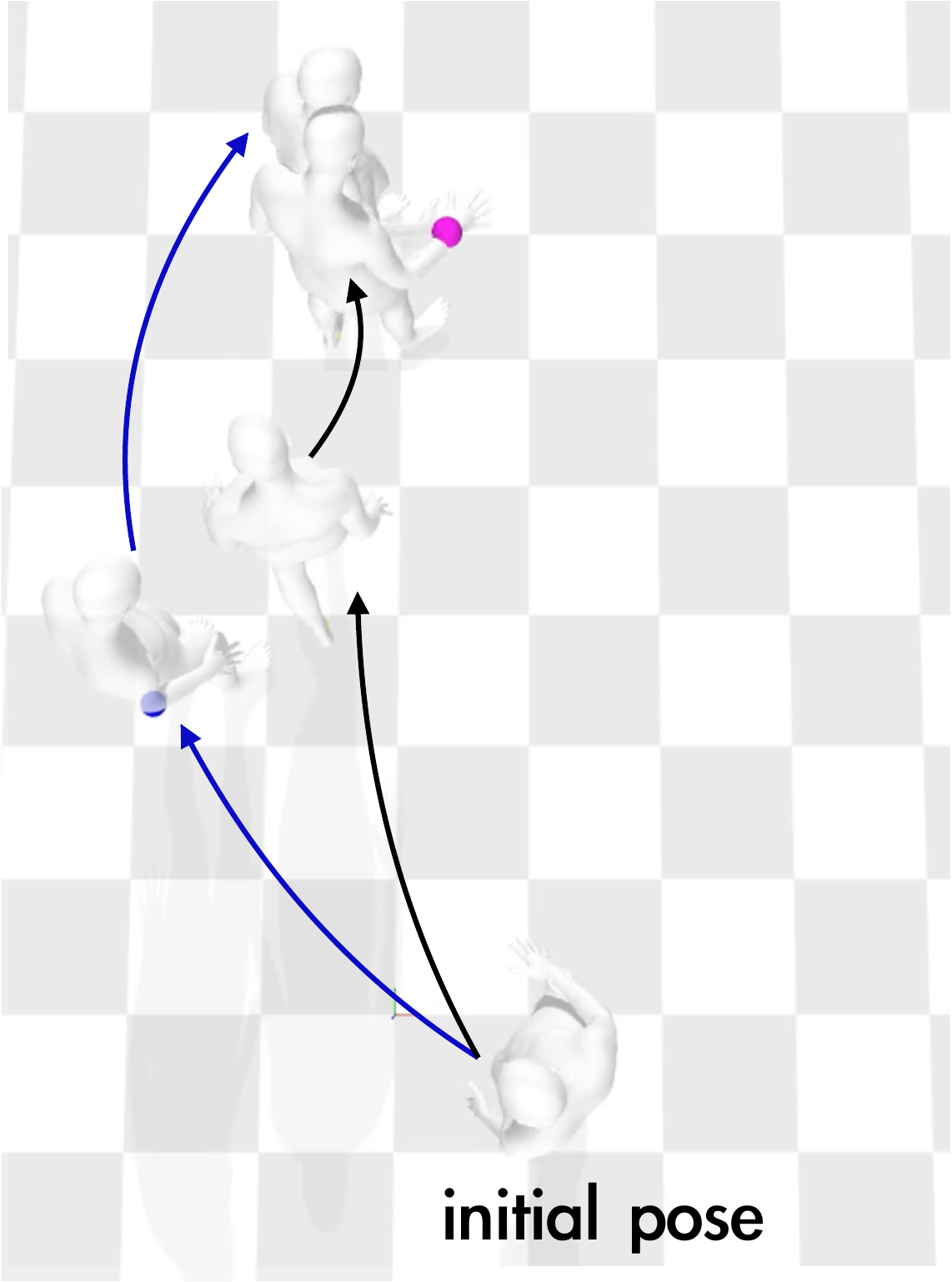}
    \caption{A generated motion from \methodname for waypoint (blue sphere) following while reaching for a goal (pink sphere). The initial generation of \methodname follows the motion marked with the black arrows. After optimization, the motion manages to pass through the waypoint, while still reaching for the goal at the end of the motion. This illustrates that \methodname provides a smooth latent space for the motions that can be aligned with the goal-reaching motion with predefined waypoints using an optimization process.}
    \label{fig:traj_following}
\end{figure}

\qheading{Waypoint Following}
Our method extends to the application of reaching goals while at the same time having the virtual human passing through arbitrary waypoints.
Specifically, we are able to choose a waypoint and have the human pass from it at a chosen frame, while still reaching for the goal at the end of the motion.
To achieve this, we extend the optimization approach described in section \cref{sec:optimization} with the addition of an extra mean square error loss between the ground projection of the pelvis and the waypoint location for a particular frame.
\cref{fig:traj_following} showcases an example of this application, underlining the adaptability of \methodname in navigating through waypoints while simultaneously reaching a goal (please see video for an example motion).
It is worth noting that passing through waypoints can be trivially extended to following trajectories, since a trajectory can be approximated by sampled waypoints on a curve.

\subsection{Extending WANDR to other joints} \label{sec:wandr_extention}

\methodname can be, in a straight forward way, extended to other joints just by replacing the wrist position with the joint of interest in the definition of $\IntentionW$. For example, by replacing the wrist with the pelvis joint, we can get a motion generator that can produce motions that follow waypoints.
However, we note that controlling multiple body joints simultaneously is not trivial with the current design. It would require redesigning the intention features to enable learning which joint corresponds to which intention feature.
\section{Perceptual Study} \label{sec:perceptual_study}
Throughout our experiments, we empirically found that the foot skating metric has a high correlation with the quality of the motion.
Nevertheless, in order to properly evaluate the perceptual quality of \methodname's generated motions we conduct two perceptual studies through amazon mechanical turk.
The studies aim at quantifying how close \methodname's motions are perceptually compared to real human motions taken from AMASS.
In the first study, users rate the realism of the motions using with a 5-level Likert scale (1 $\rightarrow$ non-realistic \& 5 $\rightarrow$ realistic). Only one motion is shown at a time.
In the second study, users are asked to choose the most realistic motion between two, one coming from \methodname and one coming from an AMASS sequence. We clip motions to a $2$ second duration and only show motions from \methodname that succeeded in reaching their goal.

In the first study, AMASS ground-truth motions score $3.8^{\pm 1}/5$ vs  $3.4^{\pm 1}/5$ for \methodname. The comparative study finds that $30.2\%$ of the users preferred \methodname motions over AMASS. These findings indicate that the \methodname motions are perceptually close to real motions.
\section{Evaluation Distribution} \label{sec:eval_distribution}
\begin{figure}
    \centering
    \includegraphics[width=0.45\textwidth, trim=0cm 0cm 0cm 0cm]{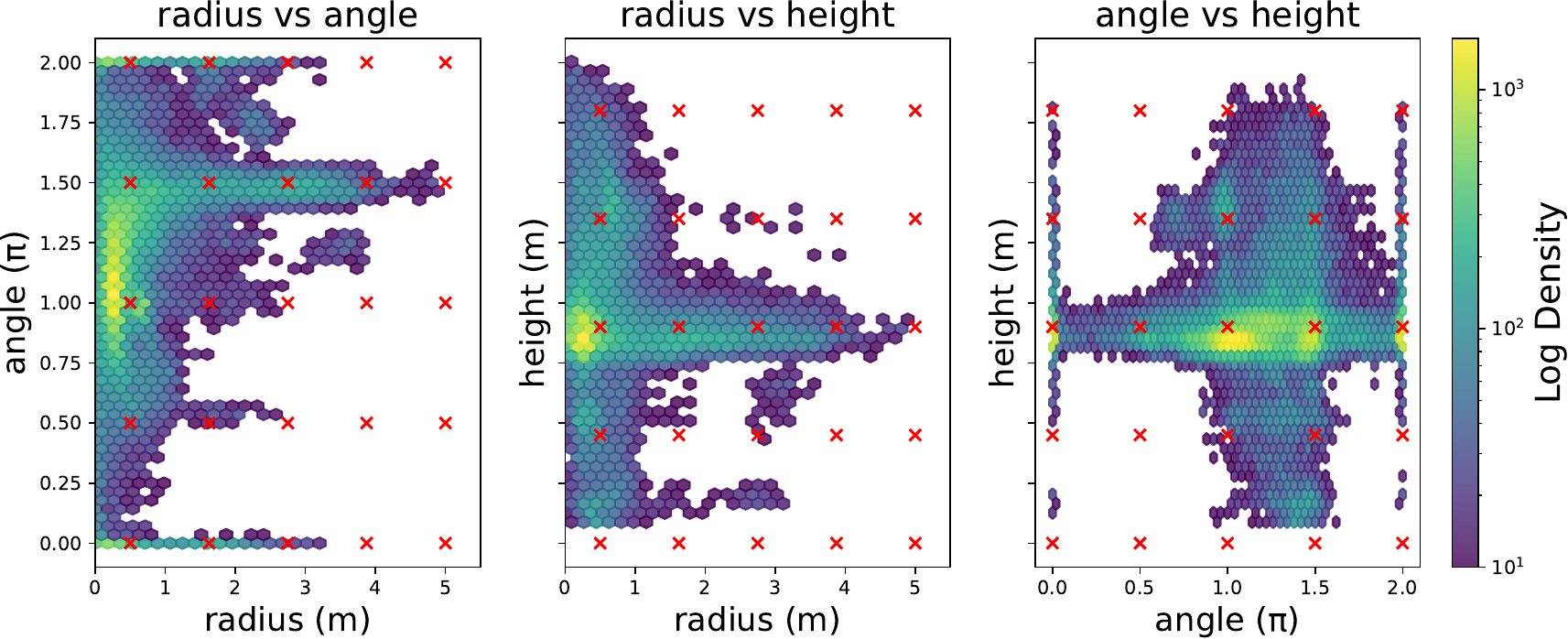}
    \caption{Overlay of the distribution of training pseudo-goals with the evaluation goals (\textcolor{red}{$\times$}) of WANDR in all pairwise combinations of their cylindrical coordinates. Our evaluation goals uniformly cover a range of goals both outside and inside the training distribution.}
    \label{fig:eval_distribution}
    \vspace{-3mm}
\end{figure}

To better demonstrate that \methodname has been evaluated on out-of-distribution data, 
in \cref{fig:eval_distribution} we visualize the density of the pseudo-goal training locations (in cylindrical coordinates) and overlay the goal locations (marked as \textcolor{red}{$\times$}) used to evaluate \methodname. We clearly observe that most evaluation goals lie on either low probability or unseen locations.

{\small
\bibliographystyle{config/ieeenat_fullname}
\bibliography{paper/references}
}

\end{document}